\documentclass[10pt,twocolumn,letterpaper]{article}

\usepackage{wacv}              % To produce the CAMERA-READY version
\usepackage{graphicx}
\usepackage{amsmath}
\usepackage{amssymb}
\usepackage{booktabs}
\usepackage{subcaption}
\usepackage{multirow}

\usepackage{array}
\usepackage[accsupp]{axessibility}

\usepackage[pagebackref,breaklinks,colorlinks]{hyperref}

\usepackage[capitalize]{cleveref}
\crefname{section}{Sec.}{Secs.}
\Crefname{section}{Section}{Sections}
\Crefname{table}{Table}{Tables}
\crefname{table}{Tab.}{Tabs.}

\def\wacvPaperID{2393} % *** Enter the WACV Paper ID here
\def\confName{WACV}
\def\confYear{2025}

\begin{document}

\title{Uncertainty-Aware Online Extrinsic Calibration:\\ A Conformal Prediction Approach}
% \author{Mathieu Cocheteux\\
% Université de technologie de Compiègne, CNRS, Heudiasyc\\
% Compiègne, France\\
% {\tt\small mathieu.cocheteux@hds.utc.fr}
% \and
% Julien Moreau\\
% Université de technologie de Compiègne, CNRS, Heudiasyc\\
% Compiègne, France\\
% {\tt\small julien.moreau@hds.utc.fr}
% \and
% Franck Davoine\\
% CNRS, INSA Lyon, UCBL, LIRIS, UMR5205\\
% Lyon, France\\
% {\tt\small franck.davoine@cnrs.fr}
% }

\author{Mathieu Cocheteux$^{1}$, Julien Moreau$^{1}$, Franck Davoine$^2$\\
{ $^1$Université de technologie de Compiègne, CNRS, Heudiasyc, France}\\
{$^2$CNRS, INSA Lyon, UCBL, LIRIS, UMR5205, France}\\
{\tt\small \{mathieu.cocheteux, julien.moreau\}@hds.utc.fr, franck.davoine@cnrs.fr}
}

\maketitle

%%%%%%%%% ABSTRACT
\begin{abstract}
Accurate sensor calibration is crucial for autonomous systems, yet its uncertainty quantification remains underexplored. We present the first approach to integrate uncertainty awareness into online extrinsic calibration, combining Monte Carlo Dropout with Conformal Prediction to generate prediction intervals with a guaranteed level of coverage. Our method proposes a framework to enhance existing calibration models with uncertainty quantification, compatible with various network architectures. Validated on KITTI (RGB Camera-LiDAR) and DSEC (Event Camera-LiDAR) datasets, we demonstrate effectiveness across different visual sensor types, measuring performance with adapted metrics to evaluate the efficiency and reliability of the intervals. By providing calibration parameters with quantifiable confidence measures, we offer insights into the reliability of calibration estimates, which can greatly improve the robustness of sensor fusion in dynamic environments and usefully serve the Computer Vision community. 
\end{abstract}
%------------------------------------------------------------------------
\section{Introduction}

In the dynamic field of autonomous systems, extrinsic calibration, which determines the spatial relationships between sensors, is essential for effective data fusion from multiple sensors. The calibration quality can thus directly impact subsequent tasks such as object detection or segmentation. In real-world scenarios, even slight calibration errors can significantly impact the safety and performance of autonomous vehicles and robots~\cite{bouain2017extrinsic}.

Traditional calibration methods, relying on manual procedures or controlled environments, are increasingly inadequate for the demands of modern autonomous systems. These approaches, while precise in controlled settings, do not allow for on-the-fly calibration, and thus to maintain a correct calibration in operation. This limitation has created a pressing need for robust techniques capable of real-time online calibration in natural environments.

Recent years have seen significant advancements in the state of the art for online calibration, particularly with the integration of deep learning-based methods. Models such as those proposed in~\cite{park2018calibnet,shiCalibRCNNCalibratingCamera2020,cocheteux2023unical,wuNetCalibNovelApproach2021,Cocheteux_2024_CVPR,jingDXQNetDifferentiableLiDARCamera2022,lvLCCNetLiDARCamera2021} have demonstrated remarkable improvements in calibration efficiency and accuracy. However, the quantification of uncertainty in the calibration process is yet to be studied. Quantifying the reliability of calibration estimates is essential in ensuring a consistent calibration quality.

In this context, we focus on model uncertainty (epistemic uncertainty), which reflects the confidence of the model in its predictions. Unlike data uncertainty (aleatoric uncertainty), which is inherent and irreducible, model uncertainty can be mitigated with more data or improved models \cite{kendall2017uncertainties}. This focus is crucial for calibration tasks, where the reliability of the model's predictions directly impacts system performance and safety.

To address this challenge, we propose an approach that integrates Monte Carlo Dropout (MCD)~\cite{galDropoutBayesianApproximation2016} with Conformal Prediction (CP)~\cite{vovk2005algorithmic}. Our method generates prediction intervals with statistically guaranteed coverage---the probability that the true outcome falls within the predicted interval---enabling robust quantification of calibration parameter uncertainty in dynamic environments. To the best of our knowledge, this work pioneers the study of uncertainty in online extrinsic calibration and introduces the first CP-based framework for providing statistically guaranteed intervals in this context. Our key contributions are:
\begin{itemize}
    \item A Conformal Prediction~\cite{vovk2005algorithmic} framework tailored for online extrinsic calibration, providing reliable and statistically sound uncertainty estimates.
    \item A deep learning-based approach that seamlessly integrates Monte Carlo Dropout \cite{galDropoutBayesianApproximation2016} for model uncertainty estimation with Conformal Prediction~\cite{vovk2005algorithmic}, enabling uncertainty quantification in dynamic environments.
    \item A comprehensive validation on real-world benchmark datasets with different sensor modalities (KITTI~\cite{geigerVisionMeetsRobotics2013} for RGB-LiDAR and DSEC~\cite{gehrigDSECStereoEvent2021} for Event Camera-LiDAR calibration), demonstrating the generalization and effectiveness of uncertainty-aware calibration across different sensor types.
\end{itemize}

\section{Related Work}
We would like to draw the attention of the reader on the two different uses of the word \textit{calibration} in this work. \textit{Sensor calibration}, or \textit{extrinsic calibration}, refers to the spatial transformation between sensors. Conversely, \textit{uncertainty calibration} refers to a process part of the CP method and described in~\Cref{sec:calibration}.

\subsection{Extrinsic Calibration for Multi-Sensor Systems}

Extrinsic calibration, the process of estimating spatial relationships between heterogeneous sensors, is crucial for accurate data fusion in autonomous driving and robotics~\cite{mirzaei3DLIDARCamera2012}. Traditional methods often relied on manual procedures or controlled environments~\cite{zhang2014extrinsic}, which proved impractical for dynamic, real-world scenarios. This limitation has driven research towards automated, robust, and online calibration techniques.

Early work in automated calibration saw significant advancements.~\cite{pandey2015automatic} introduced an automatic extrinsic calibration method for LiDAR-camera systems using mutual information maximization.~\cite{geiger2012automatic} proposed a single-shot approach for camera and range sensor calibration, leveraging checkerboards for feature extraction.

Lead by the seminal work of Schneider~\etal~\cite{schneider2017regnet}, recent years have witnessed significant advancements with the apparition of deep learning-based calibration techniques~\cite{park2018calibnet,shiCalibRCNNCalibratingCamera2020,cocheteux2023unical,wuNetCalibNovelApproach2021,Cocheteux_2024_CVPR,jingDXQNetDifferentiableLiDARCamera2022,lvLCCNetLiDARCamera2021}, demonstrating the potential of end-to-end learning in handling complex spatial relationships between sensors in uncontrolled environments. 
While these methods have improved calibration accuracy, they do not explore uncertainty quantification. Our work addresses this gap by proposing a framework to provide uncertainty estimates and safe intervals for existing calibration models. We demonstrate our approach using slightly modified versions of state-of-the-art lightweight models UniCal~\cite{cocheteux2023unical} and MuLi-Ev~\cite{Cocheteux_2024_CVPR} as case studies. This framework aims to enhance the reliability and interpretability of extrinsic calibration in dynamic, real-world scenarios.

\subsection{Uncertainty Quantification in Computer Vision}

Uncertainty quantification has become indispensable in computer vision, particularly for safety-critical applications~\cite{kendall2017uncertainties, abdarReviewUncertaintyQuantification2021}. Modern deep learning models contend with both aleatoric (data-inherent) and epistemic (model-related) uncertainties~\cite{kendall2017uncertainties}. Recent advancements have yielded diverse techniques to address these challenges.

Bayesian Neural Networks (BNNs) provide a principled approach by approximating posterior distributions over network weights~\cite{mackay1992practical}. To mitigate their computational complexity, methods like Monte Carlo Dropout have emerged, interpreting dropout as a Bayesian approximation during inference~\cite{galDropoutBayesianApproximation2016}. This enables efficient uncertainty estimation in large-scale vision applications with minimal architectural modifications.

Ensemble methods, such as Deep Ensembles~\cite{lakshminarayananSimpleScalablePredictive2017}, aggregate predictions from multiple independently trained models, effectively capturing model uncertainty while enhancing predictive performance. In medical imaging, the Probabilistic U-Net~\cite{kohl2018probabilistic} combines U-Net architecture with a conditional variational autoencoder to generate multiple plausible segmentations, addressing inherent ambiguities.

Prior Networks~\cite{malininPredictiveUncertaintyEstimation2018} explicitly model distributional uncertainty, crucial for distinguishing various uncertainty types, including out-of-distribution samples.

These methods have demonstrated efficacy across various computer vision tasks, including object detection~\cite{he2019bounding}, semantic segmentation~\cite{kendall2015bayesian}, and depth estimation~\cite{kendall2017uncertainties}. In autonomous driving, uncertainty estimates help identify low-confidence situations, potentially triggering safety interventions~\cite{feng2018towards}. However, challenges remain, including computational overhead and distributional assumptions, presenting opportunities for future research in efficient and scalable uncertainty quantification for real-world vision systems.

\subsection{Conformal Prediction (CP)}
CP has emerged as a powerful framework for uncertainty quantification, offering distribution-free guarantees for prediction intervals~\cite{vovk2005algorithmic,angelopoulosConformalPredictionGentle2023}. It has a low computational cost at runtime, and works with minimal assumptions (mostly exchangeability). Unlike traditional methods that rely on specific distributional assumptions, CP provides valid prediction sets under the minimal assumption of exchangeability, making it widely applicable across various domains~\cite{shafer2008tutorial}.

The core principle of CP lies in its ability to construct prediction intervals that contain the true outcome with a user-specified probability, regardless of the underlying data distribution~\cite{vovk2017nonparametric,toccaceliIntroductionConformalPredictors2022a}. This is achieved through a non-conformity measure, which quantifies the dissimilarity between a new example and a set of previously observed examples~\cite{toccaceliIntroductionConformalPredictors2022a}. The resulting prediction intervals adapt to the complexity of the data, offering tighter bounds in regions of high confidence and wider intervals where uncertainty is greater~\cite{angelopoulos2022image}.

Several variants of CP have been developed to enhance its efficiency and applicability. Inductive Conformal Prediction~\cite{papadopoulos2002inductive} simplifies the original framework by keeping out an uncertainty calibration set on which are computed nonconformity scores, reducing computational complexity for large datasets. Split Conformal Prediction~\cite{lei2018distribution} further refines this approach, providing a simple yet powerful method for constructing prediction intervals in regression tasks.

Recent advancements have focused on integrating CP with modern machine learning techniques. Conformalized Quantile Regression~\cite{romano2019conformalized} combines CP with quantile regression, yielding more efficient prediction intervals. The Jackknife+ method~\cite{barberPredictiveInferenceJackknife2021} offers a computationally efficient alternative that produces asymptotically valid prediction intervals under weaker assumptions.

The distribution-free nature of CP has led to its successful application in various fields, including computer vision~\cite{angelopoulos2022image}, medical diagnosis~\cite{lu2022fair}, and time series forecasting~\cite{stankeviciute2021conformal}. In computer vision, CP has been used to provide uncertainty estimates for image-to-image regression tasks~\cite{angelopoulos2022image}, demonstrating its potential for enhancing the reliability of deep learning models.

To date, the application of CP in the field of sensor calibration remains unexplored. Our work aims to integrate CP with deep learning models to develop a novel framework for online extrinsic calibration with reliable uncertainty estimates.
\section{Method}

\begin{figure}
    \centering
    \includegraphics{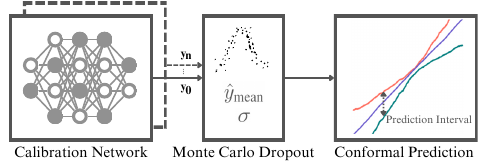}
    \caption{Overview of the uncertainty-aware calibration pipeline. (Left) The deep learning Calibration Network estimates parameters from sensor data. (Center) MCD is applied to generate multiple predictions, producing a mean estimate $\hat{y}_\text{mean}$ (the prediction) and a standard deviation $\sigma$ (measuring the uncertainty). (Right) The CP method is beforehand calibrated on a separate uncertainty calibration set, then can be used to estimate intervals with a  \(1-\alpha\) coverage.}
    \label{fig:overview}
\end{figure}

This section introduces our uncertainty-aware online extrinsic calibration approach, which is built to be used on top of a deep learning extrinsic calibration model. In this work, we conduct experiments using models based on~\cite{cocheteux2023unical,Cocheteux_2024_CVPR}. ~\Cref{fig:overview} illustrates our proposed calibration pipeline. We focus on quantifying model uncertainty (epistemic) rather than data uncertainty (aleatoric) for two key reasons: (1) model uncertainty is reducible through improved modeling and additional data, critical for enhancing calibration reliability, and (2) in our context, it is mostly linked to small variations in time synchronization, with an impact much smaller than that of the error due to the model quality. Our approach generates prediction intervals for calibration parameters with coverage guarantees, ensuring reliable calibration in dynamic environments. 

In real-world application, the extrinsic calibration parameters would then be updated in real-time using the mean prediction from MCD, and the uncertainty estimate and width interval from CP would be used to adjust the robustness of the system. If the uncertainty exceeds a certain threshold, the system could trigger a recalibration process or adjust the confidence in sensor fusion tasks.

The following subsections detail our uncertainty estimation techniques and the application of CP.

\subsection{Dropout as a Bayesian Approximation}
Bayesian Neural Networks~\cite{mackay1992practical} are good at estimating model uncertainty, but are 
often too computationally heavy for real-time applications. We thus employ a lighter method, Monte Carlo Dropout (MCD), leveraging its interpretation as an approximate Bayesian inference method \cite{galDropoutBayesianApproximation2016}. By applying dropout during both training and inference, we sample from the approximate posterior distribution of the network's weights, enabling uncertainty estimation in our calibration predictions.

Given a neural network $f(x; \theta)$ with weights $\theta$, MCD generates $N$ stochastic forward passes, each with a different dropout mask:

\begin{equation}
    \hat{y}_i = f(x; \theta_i), \quad i = 1, \dots, N
\end{equation}

where $\theta_i$ is the randomly masked weights. The mean prediction and model uncertainty are then estimated as:
\begin{equation}
    \hat{y}_{\text{mean}} = \frac{1}{N} \sum_{i=1}^{N} \hat{y}_i, \quad
    \sigma = \sqrt{\frac{1}{N} \sum_{i=1}^{N} (\hat{y}_i - \hat{y}_{\text{mean}})^2}
\end{equation}
To optimize MCD for real-time online calibration, where we process single data points sequentially, we could implement a parallel execution strategy, which consists in replicating the input $N$ times and treat it as a $N$-size batch, then perform $N$ forward passes simultaneously with different dropout masks $\{\theta_i\}$. This approach would reduce computational overhead, making it suitable for real-time applications without compromising uncertainty estimation quality.

\subsection{Building Intervals with Conformal Prediction}
To complement MCD and provide a guarantee on the prediction intervals, we integrate CP. CP is a distribution-free technique which, given a desired maximal error rate \(\alpha\) defining a coverage level \( 1 - \alpha \), produces a prediction interval that is guaranteed to contain the true calibration parameters with at least a \( 1 - \alpha \) probability.

Our method is particularly inspired by the implementation in \cite{angelopoulos2022image}, which provides a framework for generating CP intervals from a scalar uncertainty measure.

\subsubsection{Nonconformity Measure and Calibration}
\label{sec:calibration}
Our CP method integrates uncertainty estimation with rigorous statistical guarantees \cite{angelopoulosConformalPredictionGentle2023}. We define a nonconformity measure that quantifies the discrepancy between predictions and ground truth, accounting for model uncertainty \cite{shafer2008tutorial}. 

Given an uncertainty calibration set $\{(y_k^{\text{true}}, \hat{y}_k, \hat{\sigma}_k)\}_{k=1}^{m}$, where $m$ is the number of samples, $\hat{y}_k$ is the predicted calibration parameter, $y_k^{\text{true}}$ is the ground truth, and $\hat{\sigma}_k$ is the estimated uncertainty from MCD, we compute the nonconformity score $s_k$ for each sample:

\begin{equation}
    s_k = \frac{\lvert \hat{y}_k - y_k^{\text{true}} \rvert}{\hat{\sigma}_k}
\end{equation}

This score, inspired by the Mahalanobis distance \cite{mahalanobis2018generalized}, normalizes the prediction error by the estimated uncertainty, allowing for adaptive confidence intervals that account for varying levels of uncertainty across different predictions~\cite{romano2019conformalized}.

The uncertainty calibration phase involves computing these scores for the entire calibration set, resulting in $\{s_1, s_2, \dots, s_m\}$. This set forms the basis for determining the appropriate quantile used in constructing prediction intervals, ensuring that our method maintains the desired coverage level across diverse scenarios \cite{vovk2005algorithmic, lei2018distribution}.

\subsubsection{Quantile Determination}

To ensure that the prediction intervals have the desired coverage level \( 1 - \alpha\), we compute a quantile \( Q_{1-\alpha} \) from the nonconformity scores obtained in the calibration phase. Specifically, the quantile \( Q_{1-\alpha} \) is determined by sorting the nonconformity scores in ascending order \( s_{(1)} \leq s_{(2)} \leq \dots \leq s_{(m)} \) and selecting the \( (m+1)(1-\alpha) \)-th score:
\begin{equation}
    Q_{1-\alpha} = s_{\left(\lceil (m+1) \cdot (1-\alpha) \rceil\right)}
\end{equation}
where \( \lceil \cdot \rceil \) denotes the ceiling function. This selection ensures that the proportion of nonconformity scores less than or equal to \( Q_{1-\alpha} \) is at least \( 1-\alpha \), providing the desired coverage.

\subsubsection{Prediction Interval Computation}

Finally, for a new test input, we use the quantile \( Q_{1-\alpha} \) to compute the prediction interval for the extrinsic calibration parameter. Given a new prediction \( \hat{y} \) and its associated uncertainty \( \hat{\sigma} \), the prediction interval is calculated as:
\begin{equation}
    \text{Prediction Interval} = \left[\hat{y} - Q_{1-\alpha} \cdot \hat{\sigma}, \hat{y} + Q_{1-\alpha} \cdot \hat{\sigma}\right]
\end{equation}
This interval provides a guarantee that the true calibration parameter will fall within the prediction interval with at least probability \( 1-\alpha \).

By following these steps, our method leverages both the predictive power of deep learning models and the robustness of CP to provide uncertainty-aware extrinsic calibration intervals that are reliable in dynamic environments.

\subsection{Implementation Details}

Our calibration uncertainty framework is demonstrated on existing deep learning-based calibration models~\cite{cocheteux2023unical,Cocheteux_2024_CVPR}, which serve as the backbone for extrinsic calibration. The implementation of MCD requires minimal modifications to the original architecture, and CP can be treated as a post-processing step. 
Training and inference are conducted using PyTorch, and the models are evaluated on real-world datasets. The models were trained from scratch using NVIDIA V100 GPUs. To facilitate the implementation of our CP method, we utilize the Fortuna~\cite{detommaso2023fortuna} framework. 
A separate uncertainty calibration set is used to establish the nonconformity thresholds, ensuring that the prediction intervals are valid across diverse scenarios.
In line with what is done in~\cite{cocheteux2023unical,Cocheteux_2024_CVPR}, we introduce artificial decalibrations on the input during the training and testing. As we are mostly interested in the range of small decalibrations most often encountered in real-world scenarios, and requiring the best accuracy, with $\pm1^{\circ}$ and $\pm10cm$ on all axes.

\section{Experiments}
\subsection{Datasets}
We evaluate our uncertainty-aware online extrinsic calibration method on two datasets: KITTI~\cite{geigerVisionMeetsRobotics2013}, which provides synchronized RGB images and 64-channel LiDAR data, and DSEC~\cite{gehrigDSECStereoEvent2021}, which offers event camera data and 16-channel LiDAR. These datasets cover diverse sensor modalities and conditions, enabling a robust assessment of our approach.

For KITTI, we split the data into training (60\%), validation (15\%), calibration (15\%), and testing (10\%) sets. For DSEC, we use 70\% for training, 15\% for validation, and 15\% for calibration, with a separate test set. The calibration subsets are used to compute nonconformity scores for the conformal prediction method, ensuring well-calibrated uncertainty intervals.

\subsection{Evaluation Metrics}
\label{sec:metrics}
To rigorously evaluate the performance of our method, we employ a set of well-established metrics specifically tailored for CP intervals. These metrics assess both the reliability and efficiency of the prediction intervals generated by our approach.

\subsubsection{Prediction Interval Coverage Probability (PICP)}
PICP is a widely used metric~\cite{5966350,SLUIJTERMAN2024106203}, which quantifies the proportion of true calibration parameters that fall within the predicted intervals. PICP is crucial for assessing the reliability of the prediction intervals, ensuring that the true values are captured within the intervals at the desired coverage level. For our method, we calculate PICP as follows:

\begin{equation}
\text{PICP} = \frac{1}{m} \sum_{k=1}^{m} \mathbb{I} \left( y_k \in [\hat{y}_k^{\text{lower}}, \hat{y}_k^{\text{upper}}] \right)
\end{equation}

where \( y_k \) represents the true calibration parameter, \( \hat{y}_k^{\text{lower}} \) and \( \hat{y}_k^{\text{upper}} \) denote the lower and upper bounds of the predicted interval, and \( \mathbb{I}(\cdot) \) is the indicator function. A PICP close to the desired coverage level (\eg 90\%) indicates that the prediction intervals are reliable.

\subsubsection{Mean Prediction Interval Width (MPIW)}
MPIW~\cite{dewolf2023valid} measures the average width of the prediction intervals, providing insight into the trade-off between interval width and coverage. It is defined as:

\begin{equation}
    \text{MPIW} = \frac{1}{m} \sum_{k=1}^{m} (\hat{y}_k^{\text{upper}} - \hat{y}_k^{\text{lower}})
\end{equation}

While narrower intervals are generally preferred, they must still maintain the desired coverage as indicated by the PICP. MPIW helps quantify this balance, with lower MPIW values being more desirable provided that the PICP is maintained close to the target coverage level.

\subsubsection{Interval Score (IS)}
IS~\cite{gneiting2007strictly} provides a balanced evaluation by penalizing both the width of the prediction intervals and any instances where the intervals fail to cover the true value. IS is particularly useful for assessing the efficiency of the intervals, as it encourages intervals that are both narrow and reliable. The Interval Score for our method is computed as:

\begin{equation}
\begin{split}
\text{IS} = \frac{1}{m} \sum_{k=1}^{m} & \left[ (\hat{y}_k^{\text{upper}} - \hat{y}_k^{\text{lower}}) \right. \\
& + \frac{2}{\alpha} \cdot (\hat{y}_k^{\text{lower}} - y_k) \cdot \mathbb{I}(y_k < \hat{y}_k^{\text{lower}}) \\
& \left. + \frac{2}{\alpha} \cdot (y_k - \hat{y}_k^{\text{upper}}) \cdot \mathbb{I}(y_k > \hat{y}_k^{\text{upper}}) \right]
\end{split}
\end{equation}

where \( \alpha \) is the significance level (\eg 0.1 for a $1-\alpha=90\%$ coverage interval). A lower IS indicates better overall performance, reflecting more efficient and reliable intervals.

\subsection{Results}
\begin{figure}
    \centering
    \includegraphics[width=\columnwidth]{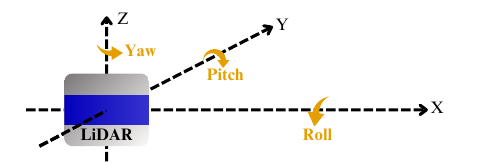}
    \caption{Axes of rotation and translation of the spatial transformation in the LiDAR frame.}
    \label{fig:axes}
\end{figure}

Throughout this subsection, results are analyzed for each axis in translation and in rotation. These axes are represented in~\Cref{fig:axes}. As we pioneer the introduction of uncertainty estimation and interval prediction for online extrinsic calibration, we will mostly assess the quality of our results in the absolute, and show that they provide added value by providing uncertainty estimates of which we will demonstrate the robustness.

\subsubsection{Interval Quality Analysis}
\label{sec:interval_quality}
\begin{table*}\resizebox{\textwidth}{!}{%
    \centering
\begin{tabular}{>{\raggedright\arraybackslash}llcccccccccccccccccc}
    \toprule
    \multirow{3}{*}{Dataset}& \multirow{3}{*}{Metric}& 
    \multicolumn{3}{c}{X (cm)}& \multicolumn{3}{c}{Y (cm)}& \multicolumn{3}{c}{Z (cm)}& \multicolumn{3}{c}{Roll ($^{\circ}$)}& \multicolumn{3}{c}{Pitch ($^{\circ}$)}& \multicolumn{3}{c}{Yaw ($^{\circ}$)}\\
    & & \multicolumn{3}{c}{Target Coverage}& \multicolumn{3}{c}{Target Coverage}& \multicolumn{3}{c}{Target Coverage}& \multicolumn{3}{c}{Target Coverage}& \multicolumn{3}{c}{Target Coverage}& \multicolumn{3}{c}{Target Coverage}\\
    & & 90\%& 95\%& 99\%& 90\%& 95\%& 99\%& 90\%& 95\%& 99\%& 90\%& 95\%& 99\%& 90\%& 95\%& 99\%& 90\%& 95\%& 99\%\\
    \midrule
    \multirow{3}{*}{KITTI~\cite{geigerVisionMeetsRobotics2013,cocheteux2023unical}}& PICP (\%) & 89.4 & 95.3 & 98.8 & 88.7 & 94.3 & 98.9 & 89.2 & 94.0 & 98.6 & 89.3 & 93.7 & 99.0 & 90.5 & 94.5 & 99.3 & 88.8 & 94.9 & 98.6 \\
    & MPIW $\downarrow$ & 2.81 & 3.62 & 5.17 & 1.64 & 2.14 & 3.42 & 2.43 & 3.10 & 4.65 & 0.14 & 0.17 & 0.26 & 0.31 & 0.39 & 0.66 & 0.15 & 0.19 & 0.29 \\
    & IS $\downarrow$ & 3.56 & 3.96 & 5.27 & 2.17 & 2.46 & 3.55 & 3.35 & 3.63 & 4.86 & 0.17 & 0.19 & 0.26 & 0.37 & 0.42 & 0.67 & 0.19 & 0.21 & 0.30 \\
    
    \midrule
  \multirow{4}{*}{DSEC~\cite{gehrigDSECStereoEvent2021,Cocheteux_2024_CVPR}} & PICP (\%) & 88.2  & 94.5 & 99.0 & 90.1  & 94.6 & 98.8 & 90.2  & 95.5 & 98.9 & 92.0  & 95.3 & 98.9 & 90.7  & 95.8 & 99.2 & 90.4  & 94.2 & 98.8\\
          & MPIW $\downarrow$ & 1.45  & 1.95 & 3.09 & 1.20  & 1.54 & 2.32 & 2.09  & 2.79 & 4.06 & 0.09  & 0.12 & 0.20 & 0.36  & 0.49 & 0.88 & 0.13  & 0.17 & 0.27\\
          & IS $\downarrow$ & 1.83  & 2.12 & 3.13 & 1.45  & 1.66 & 2.34 & 2.09  & 3.01 & 4.11 & 0.12  & 0.13 & 0.20 & 0.47  & 0.55 & 0.89 & 0.16  & 0.19 & 0.27\\
        \bottomrule
    \end{tabular}}

\caption{Evaluation metrics for our uncertainty-aware online extrinsic calibration method on KITTI~\cite{geigerVisionMeetsRobotics2013} and DSEC~\cite{gehrigDSECStereoEvent2021} datasets. We report Prediction Interval Coverage Probability (PICP), Mean Prediction Interval Width (MPIW), and Interval Score (IS) for different target coverage levels across six degrees of freedom. Lower values of IS and MPIW indicate more precise and tighter prediction intervals.}

    \label{tab:merged_results}
\end{table*}

To evaluate our method, we analyze the prediction interval quality across the KITTI and DSEC datasets. Table~\ref{tab:merged_results} shows results for three key metrics: Prediction Interval Coverage Probability (PICP), Mean Prediction Interval Width (MPIW), and Interval Score (IS) at different target coverage levels.

Our method demonstrates consistent and reliable performance, with PICP values closely matching target coverage levels across both datasets. This robustness holds despite the datasets' diverse sensor modalities and environmental conditions, highlighting the method's effectiveness in maintaining desired uncertainty bounds.

Analysis of IS and MPIW metrics reveals important insights. For translational estimates, the Y-axis consistently achieves the lowest MPIW and IS values, indicating precise lateral translation estimates crucial for accurate lane-level localization in autonomous driving.

Exceptional precision is observed in Roll and Yaw estimates, with the KITTI dataset showing 99\% coverage MPIWs of 0.26$^{\circ}$ and 0.29$^{\circ}$, and corresponding IS values of 0.26$^{\circ}$ and 0.30$^{\circ}$. These narrow intervals suggest high confidence and accuracy in rotational estimates.

However, Pitch estimation shows greater uncertainty compared to other rotational parameters, with a 99\% coverage MPIW of 0.66$^{\circ}$ (IS: 0.67$^{\circ}$) in KITTI and 0.88$^{\circ}$ (IS: 0.89$^{\circ}$) in DSEC, indicating inherent challenges likely due to sensor limitations (see Section~\ref{sec:pitch}).

Overall, these results demonstrate the method's efficacy in providing well-calibrated uncertainty estimates, balancing reliability and precision. 

\begin{figure}
    \centering
    \begin{subfigure}{\columnwidth}
        \centering
        \includegraphics[width=\columnwidth]{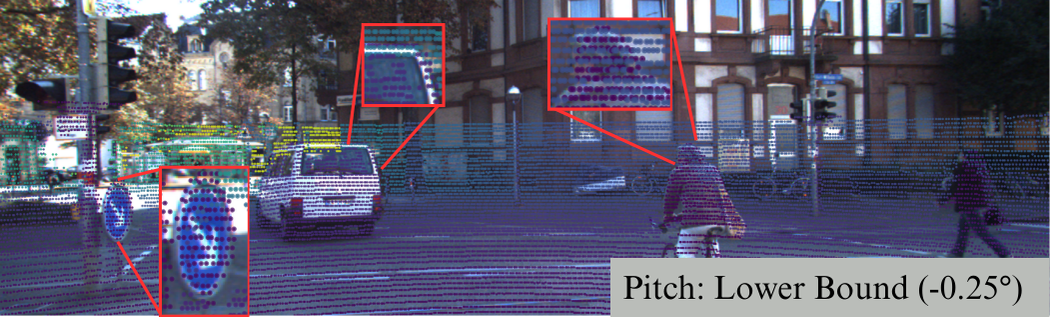}
    \end{subfigure}
    \begin{subfigure}{\columnwidth}
        \centering
        \includegraphics[width=\columnwidth]{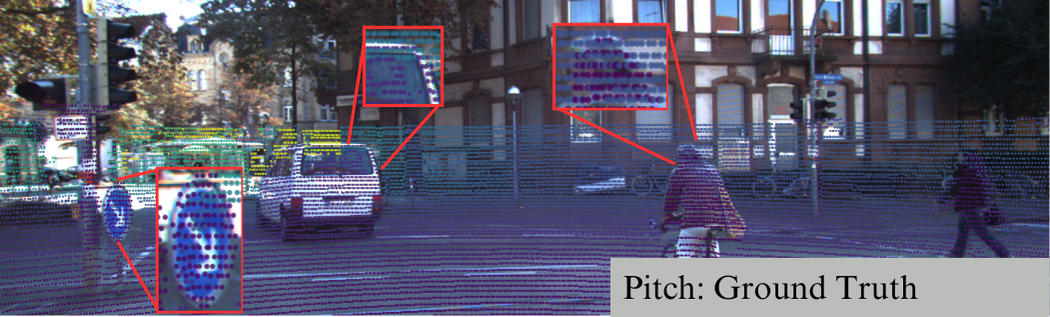}
    \end{subfigure}
    \begin{subfigure}{\columnwidth}
        \centering
        \includegraphics[width=\columnwidth]{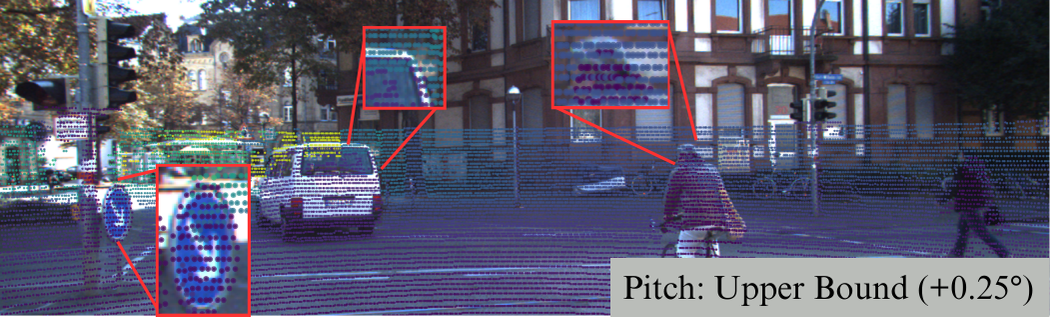}
    \end{subfigure}
    \caption{LiDAR point cloud projections onto an image frame from KITTI~\cite{geigerVisionMeetsRobotics2013} under varying Pitch calibration. (Top) Ground truth Pitch minus 0.25°. (Middle): Ground truth Pitch. (Bottom) Ground truth Pitch plus 0.25°. This ±0.25° range represents an extreme case of our 90\% confidence interval for Pitch. The visual differences in projections are minimal, demonstrating the high precision and practical robustness of our calibration method in challenging scenarios.}
    \label{fig:proj}
\end{figure}

As seen in \Cref{fig:proj}, even on the Pitch axis, and in outlier cases where the predicted 90\% interval is among the widest predicted by our network, at 0.5° (\ie ±0.25°), the difference between lower and higher bounds of the interval remains visually barely visible, as desired.

\subsubsection{Quality of the Uncertainty Calibration}
\label{sec:uncertainty_calibration}
\begin{figure}
\centering
\begin{subfigure}[b]{0.5\columnwidth}
    \centering
    \includegraphics[width=\textwidth]{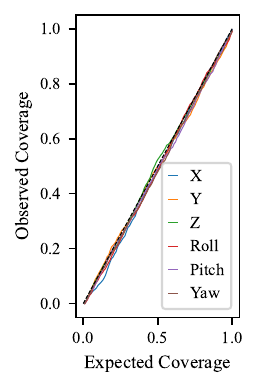}
    \caption{KITTI~\cite{geigerVisionMeetsRobotics2013}}
    \label{fig:cal_curve_kitti}
\end{subfigure}%
\begin{subfigure}[b]{0.5\columnwidth}
    \centering
    \includegraphics[width=\textwidth]{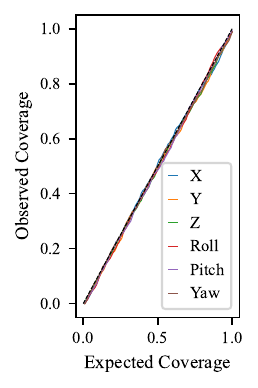}
    \caption{DSEC~\cite{gehrigDSECStereoEvent2021}}
    \label{fig:cal_curve_dsec}
\end{subfigure}
\caption{Calibration curves for extrinsic parameters on the KITTI~\cite{geigerVisionMeetsRobotics2013} and DSEC~\cite{gehrigDSECStereoEvent2021} datasets, showing observed versus expected coverage for each degree of freedom. Better seen on screen.}
\label{fig:cal_curves}
\end{figure}

To assess the calibration quality of our uncertainty estimates, we present calibration curves for both datasets in~\Cref{fig:cal_curve_kitti,fig:cal_curve_dsec}.
Both figures show the alignment of observed and expected coverage probabilities with the diagonal, indicating well-calibrated uncertainties across different coverage levels. For KITTI (Figure~\Cref{fig:cal_curve_kitti}), translational parameters (X, Y, Z) demonstrate excellent calibration, with slight deviations in some rotational parameters. The DSEC results (Figure~\Cref{fig:cal_curve_dsec}) exhibit a similar trend, confirming consistency with the PICP values in~\Cref{tab:merged_results}.

\subsubsection{Interval Visualization}
\label{sec:interval_viz}
\begin{figure*}
    \centering
    % First row of subfigures
    \begin{subfigure}{0.33\textwidth}
        \centering
        \includegraphics[width=\textwidth]{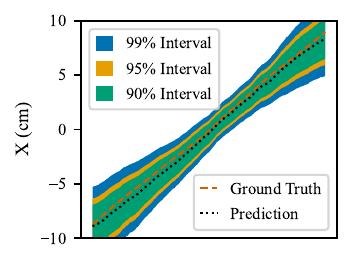}
        \label{fig:sub_x}
    \end{subfigure}
    \hfill
    \begin{subfigure}{0.33\textwidth}
        \centering
        \includegraphics[width=\textwidth]{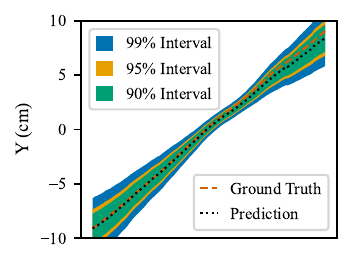}
        \label{fig:sub_y}
    \end{subfigure}
    \hfill
    \begin{subfigure}{0.33\textwidth}
        \centering
        \includegraphics[width=\textwidth]{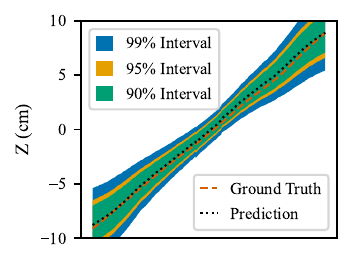}
        \label{fig:sub_z}
    \end{subfigure}
    % Second row of subfigures
    \begin{subfigure}{0.33\textwidth}
        \centering
        \includegraphics[width=\textwidth]{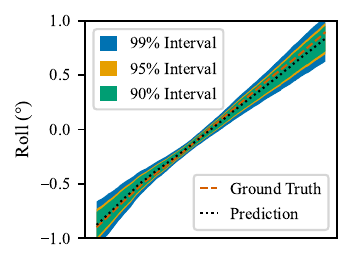}
        \label{fig:sub_roll}
    \end{subfigure}
    \hfill
    \begin{subfigure}{0.33\textwidth}
        \centering
        \includegraphics[width=\textwidth]{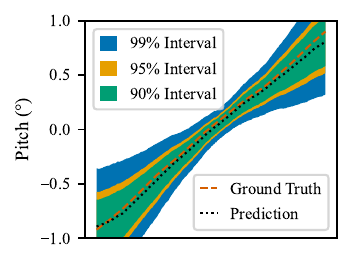}
        \label{fig:sub_pitch}
    \end{subfigure}
    \hfill
    \begin{subfigure}{0.33\textwidth}
        \centering
        \includegraphics[width=\textwidth]{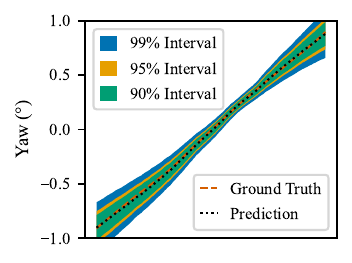}
        \label{fig:sub_yaw}
    \end{subfigure}
    \caption{Ordered prediction interval plots for the six degrees of freedom (in translation and rotation) on the KITTI~\cite{geigerVisionMeetsRobotics2013} dataset. The overlaid shaded regions represent the intervals corresponding to expected coverage levels of 90\%, 95\%, and 99\%. The ground truth values should fall within the respective intervals for at least the specified proportion of samples. The X-axis represents the ordered test samples sorted by the ground truth values for each degree of freedom, while the Y-axis indicates the deviation from the ground truth. All curves are smoothed using a moving average to enhance readability.
}
    \label{fig:ordered_intervals_kitti}
\end{figure*}

To provide a more intuitive understanding of our method's performance, we present ordered interval plots for the KITTI dataset in~\Cref{fig:ordered_intervals_kitti}. Those for DSEC can be found in the supplementary materials. These plots offer a visual representation of the prediction intervals and their relationship to the ground truth values for each degree of freedom.

We observe that the prediction intervals consistently envelop the ground truth values. This visual confirmation aligns with the PICP values reported in ~\Cref{tab:merged_results} close to the target coverage. The intervals appear to widen at the extremes of the value range, indicating increased uncertainty in these regions. This increased uncertainty is especially visible on the Pitch interval curves, which are much wider. This behavior demonstrates our method's ability to adapt its uncertainty estimates based on the difficulty of the calibration task in different scenarios.

\subsection{Discussion and Ablation Studies}
The results discussed above demonstrate the effectiveness of our method across different datasets and sensor modalities. It is important to mention that these results are reached without deteriorating the original accuracy of the calibration model. For example, we achieved on KITTI an average rotation error of $0.04^{\circ}$ and a translation error of $0.46\,\text{cm}$, comparable to~\cite{cocheteux2023unical} (rotation errors between $0.03^{\circ}$ and $0.04^{\circ}$, and translation errors from $0.33\,\text{cm}$ to $0.89\,\text{cm}$).
To reflect on these results and their robustness, we conducted ablations studies and discuss our method's strengths and weaknesses below.

\subsubsection{Impact of Monte Carlo Dropout Parameters}

\begin{figure}
    \centering
    \includegraphics[width=\columnwidth]{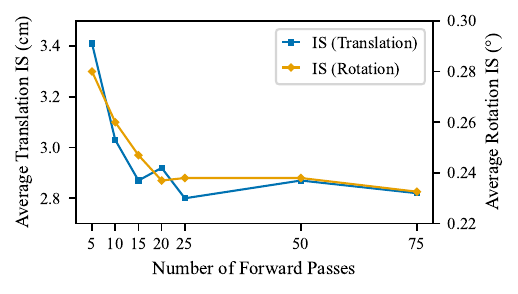}
    \caption{Impact of the number of Monte Carlo Dropout forward passes on the Interval Score (IS) averaged for translation parameters and rotation parameters. Experiment realized on KITTI~\cite{geigerVisionMeetsRobotics2013}.}
    \label{fig:mcd_ablation}
\end{figure}

We conducted an ablation study to investigate the effect of varying the number of forward passes in MCD on uncertainty estimation. \Cref{fig:mcd_ablation} illustrates the relationship between the number of forward passes and IS for both translation and rotation errors.

The results reveal a clear trend: as the number of forward passes increases, IS decreases, indicating more precise uncertainty estimates. This trend is particularly pronounced in the range of 5 to 25 forward passes for both translation and rotation errors. Beyond 25 passes, the rate of improvement diminishes significantly, suggesting a point of diminishing returns. For an optimal balance between computational efficiency and uncertainty estimation quality, we selected 25.

As demonstrated in~\cite{verdojaNotesBehaviorMC2021}, the optimal dropout rate for estimating epistemic uncertainty depends on the network architecture and its size. Following the grid search approach outlined in~\cite{galDropoutBayesianApproximation2016,galConcreteDropout2017}, we tested rates between 0.05 and 0.5. We found that the smallest rate yielding consistent results was optimal, as increasing the rate degraded performance, though with a lesser impact than the number of forward passes. Consequently, a dropout rate of 0.25 was applied to the backbone layers, while the best results were achieved with 0.05 on KITTI and 0.10 on DSEC for the head.

\subsubsection{Necessity of Applying Conformal Prediction}

\begin{table}
\centering
\begin{tabular}{ccccccc}
\toprule
Target & \multicolumn{6}{c}{PICP (\%)} \\ 
(\%) & X & Y & Z & Roll & Pitch & Yaw \\
\midrule
90 & 77.3 & 90.3 & 76.9 & 79.69 & 85.0 & 80.6 \\
95 & 82.3 & 92.0 & 81.1 & 83.85 & 88.9 & 85.8 \\
\bottomrule
\end{tabular}%

\caption{Results of the ablation study on KITTI~\cite{geigerVisionMeetsRobotics2013} showing the observed coverage (PICP) when using only MCD with a normal distribution assumption, instead of our proposed MCD+CP method.}
\label{tab:mcd_normal_results}
\end{table}

To evaluate the effectiveness of our CP approach, we performed an ablation study on KITTI using MCD \cite{galDropoutBayesianApproximation2016} alone, with a normal distribution assumption for interval estimation. In this setup, MCD was applied during inference by performing multiple forward passes with dropout enabled. The mean and standard deviation of these predictions were then calculated to estimate quantiles under the normal distribution. This method differs from our primary approach, where $\sigma$ is directly incorporated into the CP framework.

As shown in \Cref{tab:mcd_normal_results}, the MCD with a normal distribution assumption consistently underestimates interval widths, resulting in lower-than-expected coverage rates at both 90\% and 95\% target levels. This underperformance likely arises from the inadequacy of the normal distribution in capturing true uncertainty and the inherent limitations of MCD in providing calibrated uncertainty estimates for this task. These findings highlight the superiority of our CP-based method for generating more accurate and reliable prediction intervals in extrinsic calibration.

\subsubsection{Impact of LiDAR Vertical Resolution}
\label{sec:pitch}

Our experiments reveal a significant pattern in the uncertainty estimates, particularly for the Pitch angle. We observe consistently wider prediction intervals for Pitch compared to Roll and Yaw, which we attribute to the inherent limitations of LiDAR vertical resolution. This sparse vertical sampling introduces challenges in precisely aligning LiDAR and camera data along the vertical axis.

This phenomenon is especially pronounced in the DSEC dataset, which utilizes a 16-channel LiDAR, offering substantially lower vertical resolution than the 64-channel LiDAR employed in KITTI. As evidenced in~\Cref{tab:merged_results}, the MPIW for Pitch (up to 0.88°) in DSEC significantly exceeds those for Roll (0.20°) and Yaw (0.27°).
This underscores the importance of comprehensive uncertainty quantification in extrinsic calibration, especially when dealing with sensors that have inherent resolution limitations.

\subsubsection{Generalization and Practical Implications}

While we observe slight variations in performance between the KITTI and DSEC datasets, these differences are relatively minor. Overall, our method demonstrates strong generalization capabilities.

The consistency of the uncertainty estimates demonstrated in~\Cref{sec:uncertainty_calibration} and the tightness of the intervals shown in~\Cref{sec:interval_quality,sec:interval_viz} are noteworthy. These results suggest that our method  provides reliable and actionable uncertainty estimates and prediction intervals for autonomous driving systems. Its integration in calibration systems could potentially improve safety and calibration-related decision-making in real-world deployments. The method's ability to adapt to sensor limitations, such as poor vertical resolution in LiDARs, makes it particularly valuable for robust calibration in diverse autonomous driving scenarios.
\section{Conclusion}

We introduced a novel approach to online extrinsic calibration that incorporates rigorous uncertainty quantification through a combination of Monte Carlo Dropout (MCD) and Conformal Prediction (CP). MCD captures model uncertainty by providing a probabilistic measure of calibration parameters, while CP offers prediction intervals with formal guarantees, ensuring coverage of the true calibration parameters with a user-specified probability \(1 - \alpha\).

Our method enhances existing calibration models by adding the capacity to evaluate the uncertainty of its estimates and provide statistically-guaranteed intervals. The results show that our approach maintains calibration accuracy while offering insights into the reliability of these estimates, which is vital for robust sensor fusion in autonomous systems.

Future work could explore the integration of this framework in real-world systems and optimize the use of uncertainty measures to ensure consistent calibration quality.

{\small
\bibliographystyle{ieee_fullname}
\bibliography{egbib}
}

\end{document}